\documentclass{svproc}
\usepackage{url}

\usepackage{amsfonts,amsmath,amssymb}
\usepackage{amstext}

\usepackage{algorithm}
\usepackage{algorithmic}
\usepackage{graphicx}

\begin{document}
\mainmatter              % start of a contribution
\title{Weak Relation Enforcement for Kinematic-Informed Long-Term Stock Prediction with Artificial Neural Networks}
\titlerunning{Weak Relation Enforcement for Kinematic-Informed ANN}  % abbreviated title (for running head)
\author{Stanislav Selitskiy
%Ivar Ekeland\inst{1} \and Roger Temam\inst{2}
%Jeffrey Dean \and David Grove \and Craig Chambers \and Kim~B.~Bruce \and
%Elsa Bertino
}
\authorrunning{Stanislav Selitskiy
%Ivar Ekeland et al.
} % abbreviated author list (for running head)
\tocauthor{Stanislav Selitskiy
%Ivar Ekeland, Roger Temam, Jeffrey Dean, David Grove,
%Craig Chambers, Kim B. Bruce, and Elisa Bertino
}
\institute{
%Princeton University, Princeton NJ 08544, USA,
University of Bedfordshire, Park Square, Luton, LU1 3JU, UK\\
\email{
%I.Ekeland@princeton.edu
stanislav.selitskiy@study.beds.ac.uk}
%,
%\\ WWW home page:
%\texttt{http://users/\homedir iekeland/web/welcome.html}
%\and
%Universit\'{e} de Paris-Sud,
%Laboratoire d'Analyse Num\'{e}rique, B\^{a}timent 425,\\
%F-91405 Orsay Cedex, France
}

\maketitle              % typeset the title of the contribution

\begin{abstract}
We propose loss function week enforcement of the velocity relations between time-series points in the Kinematic-Informed artificial Neural Networks (KINN) for long-term stock prediction. Problems of the series volatility, Out-of-Distribution (OOD) test data, and outliers in training data are addressed by (Artificial Neural Networks) ANN's learning not only future points prediction but also by learning velocity relations between the points, such a way as avoiding unrealistic spurious predictions. The presented loss function penalizes not only errors between predictions and supervised label data, but also errors between the next point prediction and the previous point plus velocity prediction. The loss function is tested on the multiple popular and exotic AR ANN architectures, and around fifteen years of Dow Jones function demonstrated statistically meaningful improvement across the normalization-sensitive activation functions prone to spurious behaviour in the OOD data conditions.
Results show that such architecture addresses the issue of the normalization in the auto-regressive models that break the data topology by weakly enforcing the data neighbourhood proximity (relation) preservation during the ANN transformation.
\end{abstract}

\section{Introduction}
The stock forecast problems are rarely considered either learning on graphs or physics-informed learning problems; therefore, the paper title explanation is due. Unlike in the mathematical function definition, in which only one entity from the range can be associated with each entity from the domain of the function (though multiple domain entities may be associated with the range entity), mathematical relations, which the graph structures depict, do not have such limitations: multiple entities from domain may be associated with multiple entities in the range \cite{velivckovic2017graph,frasconi1998general,sperduti1997supervised}. 
Relations between the past financial series points with multiple future points require using the graph structures directly or indirectly. Such implicit relations can be provided by Artificial Neural Networks (ANN) training; however, artefacts in the training data or Out-of-Distribution (OOD) test data may generate unrealistic relations between the data points. In terms of the time series, that would be highly improbable prediction jumps in a short period of time. 

However, it's possible to explicitly enforce or hint ANN to satisfy particular relations if we know them ahead, even without statistical learning. If we know the first and maybe second derivatives over time of the time series change, and if we postulate that particular time series has 'inertia', then the real future predictions could be expected to be found in the vicinity of the velocity and acceleration-based analytical solution. The objective of this research is to reduce the spurious prediction problem of the pure ANN approach, utilizing velocity hints of the temporal graph approach through the loss function.

The nature of such relations we plan to use, i.e. velocity and acceleration (first and second derivatives of the index value with respect to time) in the broad sense, are categories of physics. Therefore, we merge two domains of Machine Learning (ML) here: Learning on Graphs (LoG) and Physics-informed (or Physics-aware) Learning \cite{RAISSI2019686,cai2022physics}, in their more narrow ANN context of Graph Neural Networks (GNN) and Physics-Informed Neural Networks (PINN). Because we do not include an analysis of the moving forces, a more narrow branch of Physics - Kinematics \cite{meyer2022robust} - is more appropriate here: Kinematics-informed Neural Networks (KINN).

The LoG domain started in the late 90s with attempts to utilise feed-forward or Recursive Neural Networks (RNN) to encode Directed Acyclic Graph (DAG) representations of the structured data as a sequence of state transitions representing edges and RNN's hidden states representing nodes \cite{sperduti1997supervised,frasconi1998general}. Later, a dedicated Graph Neural Network (GNN) was introduced to break through the DAG structure limitations and represent generic graph structures \cite{gori2005new,scarselli2008graph}. GNN of various flavours may concentrate on different points of interest: node or relations/vertices classification, graph generation, etc. In our straightforward approach, the main point of interest is node prediction, with relations having only a supportive role for that, utilizing the KINN approach. Attention mechanism became the bleeding area of ANN research in the last few years \cite{bahdanau2014neural,luong2015effective}, including GNNs \cite{velivckovic2017graph}. However, the dot-product attention mechanism is prone to spurious behaviour at close angles and OOD conditions, which may lead to failures of foundation models \cite{Blodgett2021,Bommasani2021,bender-koller-2020-climbing,Marcus2018,Lake2020}. The dot product attention mechanism is not the only one possible to implement; however, alternative ones investigated here are also prone to spurious behaviour, and experiments using KINN to improve that behaviour are presented for those models in addition to the Transformer type.

The contribution is organized in the following way: Section~\ref{sec:ps} presents the proposed solution,  Section~\ref{sec:ds} describes data sets and their partition, which were used in computational experiments, as well as accuracy metrics for algorithms' evaluation. Section~\ref{sec:ex} lists hardware parameters and model configurable parameters, as well as details of the less-popular ANN architectures implemented from scratch. Section~\ref{sec:rs} shows the results of the experiments in the diagram and table form, and Section~\ref{sec:con} discusses these results, draws conclusions and outlines future research directions.

\section{Proposed Solution}
\label{sec:ps}
GNNs are typically represented in a general form as an ordered pair of node space $\mathcal{V}$ endowed with a space of relations or vertices 
$\mathcal{E}$ between nodes $v \in \mathcal{V}$: $\mathcal{G}=(\mathcal{V}, \mathcal{E})$. Where $\forall e \in \mathcal{E}$, $e$ is an ordered pair: $e=(v, \, v^{'}) \in \mathcal{V} \times \mathcal{V}$. In our case, where entity $v$ represents stock index, $v \in \mathcal{V} \subset \mathbb{R}$ and $e_{t_{12}}=(v_{t_1}, \, v_{t_2}) = v_{t_2} - v_{t_1}$, where $t_1$ and $t_2$ are subsequent moments in time such that $t_2 - t_1 = 1$, and $e$ represent the rate of change or velocity of the stock index. 

If we represent our GNN as a transformation $gnn$ from a domain space $\mathcal{X}$ of vectors representing $T_p$ stock indexes in the past $\textbf{v} \in \mathcal{V} \subset \mathbb{R}^{T_p}$ and their velocities $\textbf{e} \in \mathcal{E} \subset \mathbb{R}^{T_p-1}$, into a range space $\mathcal{Y}$ of the future representing $T_f$ stock index positions and $T_f-1$ their velocities, and, joining spaces of indexes and their velocities, we get:

\begin{equation}
\begin{split}
gnn:\mathcal{X} &= \mathcal{V} \bigoplus \mathcal{E} \subset \mathbb{R}^{2T_p-1} \mapsto \\ & \mapsto \mathcal{Y} = \mathcal{V} \bigoplus \mathcal{E} \subset \mathbb{R}^{2T_f-1}
\end{split}
\label{eq:1}
\end{equation}

The loss function to keep spurious index predictions in check consists of two parts - the traditional sum of squares of deltas between predictions $v_{t}$ and observed values $\hat{v_{t}}$ at time steps $t$, and a sum of squares of deltas between the stock index value on the next time step and previous time step plus velocity $e_{t-1}$ of the index:

\begin{equation}
\begin{split}
L = L_v + L_ve &= \frac{\sum_{t=1}^{T_f}(v_{t} - \hat{v_{t}})^2}{T_f} + \\ &\frac{\sum_{t_i=2}^{T_f}(v_{t} - (v_{t-1}+e_{t-1}))^2}{T_f-1}
\end{split}
\label{eq:2}
\end{equation}

\section{Data sets}
\label{sec:ds}
\begin{figure}
\begin{minipage}[b]{1.0\linewidth}
  \centering
  \fbox{\rule[0cm]{0cm}{0cm}
  \centerline{
  \includegraphics[width=0.99\linewidth]{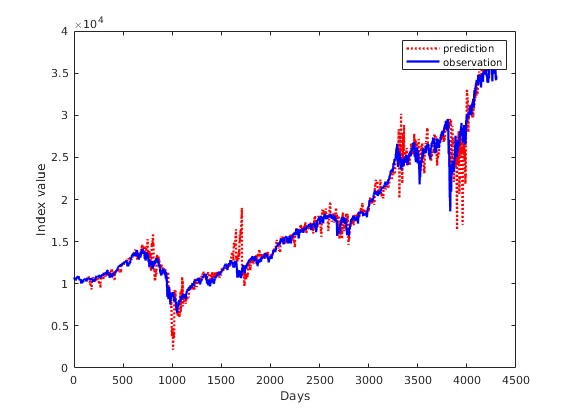}
  }
  \rule[0cm]{0cm}{0cm}}
\end{minipage}
\begin{minipage}[b]{1.0\linewidth}
  \centering
  \fbox{\rule[0cm]{0cm}{0cm}
  \centerline{
  \includegraphics[width=0.99\linewidth]{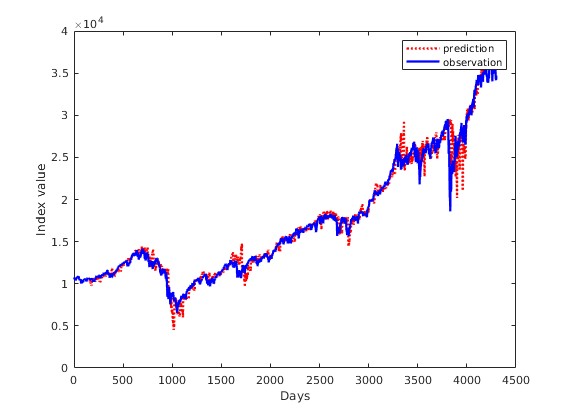}
  }
  \rule[0cm]{0cm}{0cm}}
\end{minipage}

\caption{Dow Jones predictions for KGate model, non-kinematic informed ANN (top), and kinematic informed KIANN (bottom). Normalized input}
\label{fig:dj.n}
\end{figure}

\begin{figure}
\begin{minipage}[b]{1.0\linewidth}
  \centering
\fbox{\rule[0cm]{0cm}{0cm}
  \centerline{
  \includegraphics[width=0.99\linewidth]{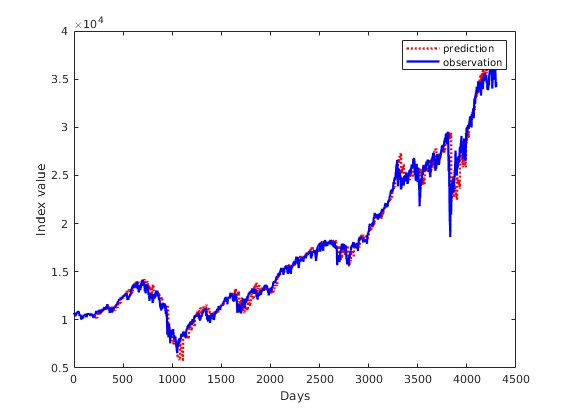}
  }
  \rule[0cm]{0cm}{0cm}}
%  \centerline{b. Non-normalized input}\medskip
\end{minipage}
\begin{minipage}[b]{1.0\linewidth}
  \centering
\fbox{\rule[0cm]{0cm}{0cm}
  \centerline{
  \includegraphics[width=0.95\linewidth]{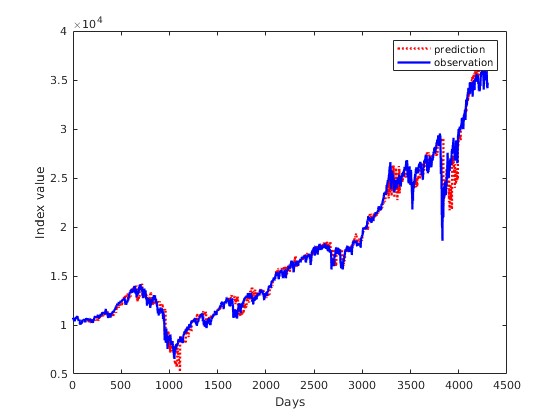}
  }
  \rule[0cm]{0cm}{0cm}}
%  \centerline{b. Non-normalized input}\medskip
\end{minipage}

\caption{Dow Jones predictions for KGate model, non-kinematic informed ANN (top), and kinematic informed KIANN (bottom). Non-normalized input}
\label{fig:dj.nn}
\end{figure}

In the computational experiments conducted, data for NASDAQ, Dow Jones, NIKKEI, and DAX indices were utilized, spanning from the beginning of the year $2005$ to the end of January $2022$. The data were retrieved from the source: \url{https://tradingeconomics.com}. It is noteworthy that a comprehensive statistical analysis of the dataset falls beyond the purview of this paper. Nevertheless, general intuition for these non-stationarity time series can be gained from Figures~\ref{fig:dj.n}-\ref{fig:dj.nn} where Dow Jones index prediction is shown for the normalized KGate models in their non-kinematic informed and kinematic-informed variants.
The dataset includes various behavioural patterns, encompassing oscillations in stagnant markets, explosive growth and declines during bubble bursts and busts across American, European, and Asian markets. The slight asynchrony in the data, amounting to a few days, is attributed to disparities in the working day schedules of various stock exchanges.

\begin{figure}
\centering
\fbox{\rule[0cm]{0cm}{0cm}
  \includegraphics[width=1.0\linewidth]{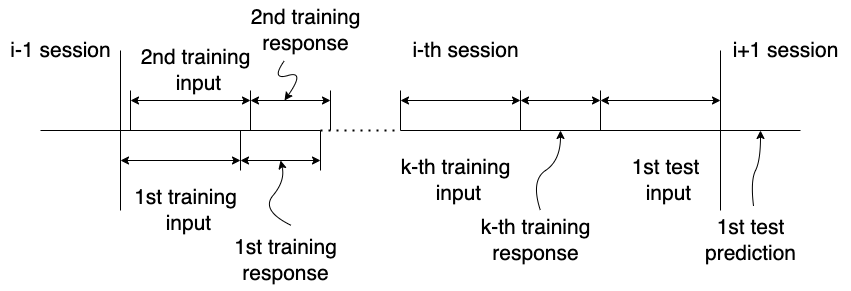}
\rule[0cm]{0cm}{0cm}}
\caption{Session partition schema.}
\label{fig:sess.part}
\end{figure}

To facilitate the long-term prediction spanning $30$ days into the future, based on the preceding $30$ days' performance of the indices, the dataset was partitioned into $35$ subsets. Each subset comprised $30$ observations, serving as training data for the sequence-to-sequence models. Each ``observation" encompassed values from a $30$-day period, with the corresponding ``label" values representing days $31$ through $60$. Subsequent observations initiated with a one-day shift. The total number of training sessions equalled $34$, and the duration of each session was set at $120$ days. This configuration ensured that none of the training data, including the label days, overlapped with the subsequent test session data. The number of test sessions amounted to the last $34$ sessions. The final $30$ days from the preceding training session, not utilized in the training process, were employed to predict the initial 30 days of the subsequent test session with a step size of $1$ until the conclusion of the test session. Model parameters were reset for each session, initiating a fresh training process (see Figure~\ref{fig:sess.part}). For sequence-to-value models, such as LSTM, the entire last label day, encompassing $119 + 1$ days, was incorporated in the training session.

As an accuracy metrics, we use the Mean Absolute Percentage Error (MAPE), defined as follows:
\begin{equation}
MAPE=\frac{1}{n}\sum_{t=1}^n |\frac{A_t-F_t}{A_t}|
\end{equation}
And Root Mean Square Error (RMSE): 
\begin{equation}
RMSE=(\frac{1}{n}\sum_{t=1}^n{(A_t-F_t)^2})^{\frac{1}{2}}
\end{equation}

Where $A_t$ and $F_t$ are the actual and predicted indexes at a given day $t$, respectively, and $n$ is the number of test observations.

\section{Experiments}
\label{sec:ex}
The experiments were run on the Linux (Ubuntu 20.04.3 LTS) operating system with two dual Tesla K80 GPUs (with $2\times 12$GB GDDR5 memory each) and one QuadroPro K6000 (with $12$GB GDDR5 memory, as well), X299 chipset motherboard, 256 GB DDR4 RAM, and i9-10900X CPU. Experiments were run using MATLAB 2022a with Deep Learning Toolbox. 

The experimentation encompassed a variety of models, including Linear Artificial Neural Network Regression (without activation functions), ANN with Rectified Linear Unit (ReLU), Logistic, and Hyperbolic Tangent activations, as well as Long Short-Term Memory (LSTM) and Gated Recurrent Unit (GRU) architectures. Additionally, simple sequential and spectral cascade Convolutional Neural Networks (CNNs), Radial Basis Function (RBF), KGate, Group Method of Data Handling (GMDH), and Transformer-type ANNs with dot-product non-linearity \cite{Selitskiy2023Batch} were explored.
Given that ANN regression, ANN with ReLU, KGate activations, and CNNs exhibit tolerance to non-normalized input data and often yield enhanced accuracy, experiments for these models were conducted using non-normalized input. Conversely, for other models sensitive to normalization, a strict min-max normalization was applied. This normalization was calculated solely for a given observation data, without considering information from either the past or future.

For KGate, which uses saturate-able logistic and hyperbolic tangent gate mechanisms, a normalized version was also run. The mean squared error was used as a loss function for all ANNs. For each standard ANN regression version, a kinematic-informed (KINN) model version was implemented and run. The above-described by the Formula~\ref{eq:2} loss function was implemented in the custom regression layer. Also, the needed velocity values were added to the input training and test data during the preprocessing step, almost doubling the ANN width, including hidden layers. 

Details of well-established ANN layers and architectures, readily accessible in MATLAB or its Toolboxes, are not extensively discussed here. Only non-default configuration parameters are provided below. For non-standard implementations and the complete source code employed in the experiments, they can be accessed on GitHub (\url{https://github.com/Selitskiy/LTTS}), where the implementations are documented as follows.

If we look at linear regression, or the linear part of a layer transformation of an ANN, as a transformation from a higher dimensional space into the lower dimensional space $f, \, n<m$:

\begin{equation}
f:\mathcal{X}\subset \mathbb{R}^m \mapsto \mathcal{Y} \subset \mathbb{R}^n
\label{eq:3}
\end{equation}

Linear regression can be represented as a matrix multiplication:

\begin{equation}
\textbf{y} = f(\textbf{x}) = \textsf{W}\textbf{x}, \, \forall \textbf{x} \in \mathcal{X}\subset \mathbb{R}^m, \, \forall \textbf{y} \in \mathcal{Y} \subset \mathbb{R}^n
\label{eq:4}
\end{equation}
where $\textsf{W} \in \mathcal{W} \subset \mathbb{R}^{n \times m}$ is the adjustable coefficient matrix, which, using minimization of the sum of squared errors, can be found as \cite{hastie01statisticallearning}:

\begin{equation}
\textbf{W} = (\hat{\textbf{X}}^T \hat{\textbf{X})}^{-1} \hat{\textbf{X}}^T \hat{\textbf{Y}}
\label{eq:5}
\end{equation}
Where $\hat{\textbf{X}}, \, \hat{\textbf{Y}}$ are matrices of the observations of the input and output observations, respectively.

To ensure a comparable level of complexity, all Artificial Neural Network (ANN) models were configured with two hidden layers. The number of neurons in the first and second hidden layers was set to $m$ and $2m+1$, respectively, where $m$ represents the input dimensionality, as expressed in Formula~\ref{eq:3}.

The reason for that was if one looks at ANN as a Universal Approximation, according to the Kolmogorov-Arnold superposition theorem \cite{kolmogorov1961representation}, for general emulation of the $f:\mathcal{X}\subset \mathbb{R}^m \mapsto \mathcal{Y} \subset \mathbb{R}$ process, that is a minimal ANN configuration needed (given that activation functions are complex enough):

\begin{equation}
\label{eq:6}
f(\textbf{x}) = f(x_1, \dots , x_m) = \sum_{q=0}^{2m} \Phi_q (\sum_{p=1}^{m} \phi_{qp}(x_p))
\end{equation}
where $\Phi_q$ and $\phi_{qp}$ are continuous $\mathbb{R} \mapsto \mathbb{R}$ functions.

The applicability of ANNs as Universal Approximators has been challenged, as discussed in \cite{girosi1989representation}, primarily due to the non-smooth nature of the inner $\phi_{qp}$ functions, rendering them impractical. Nevertheless, these concerns were refuted in \cite{Kurkova1991}. In \cite{pinkus1999approximation}, the $\phi_{qp}$ activation functions are called as "pathological". 
Consequently, in exploring alternative options, beyond the conventional activation functions, less commonly employed architectures and activation functions were subjected to experimentation.

For example, Radial Basis Functions (RBF) were introduced into ANNs dates back to the late 1980s, as proposed by Broomhead and Lowe in \cite{broomhead1988radial}. This approach can be conceptualized as a ``soft gate", wherein the activation of transformation matrix coefficients follows a Gaussian distribution proportional to the proximity of the test signal to the training signals at which the transformation matrix coefficients were trained, as discussed in \cite{Park1991}.

\begin{equation}
\label{eq:7}
f(\textbf{x}) = \textbf{a}_k e^{-\textbf{b}_k(\textbf{x}-\textbf{c}_k)^2}
\end{equation}
where $k \in \{1 \dots n\}$.

An evident limitation of this architecture lies in its ``fluffiness", arising from the non-reuse of neurons for ``missed" test-time data inputs. This characteristic renders RBF ANNs less dense or compact compared to ReLU ANNs. Despite this, RBF remains a viable architecture, finding application in niche domains as documented in \cite{kurkin2018artificial,beheim2004new}.

Another variant of the Artificial Neural Network (ANN) architecture, as detailed in \cite{Selitskiy2022}, can be categorized within the Gated Linear Unit (GLU) family of activations. This architecture employs a Directed Acyclic Graph (DAG) ANN, allowing for the implementation of a cell (referred to as Kolmogorov's Gate or KGate for brevity) comprising perceptrons with logistic sigmoid activations. These perceptrons function as allow or do not allow gates in the saturation domain, or they serve as multiplicative scaling factors for the main trunk of the ANN in the non-saturation domain of input values. Additionally, perceptrons with hyperbolic tangent activation operate as update/forget or mean shift gates on the main ANN trunk, collaborating with linear input transformation through the multiplication gate, as defined in Formula~\ref{eq:8}.

\begin{equation}
\label{eq:8}
\begin{split}
\textbf{z}_i &= (\textsf{W}_i\textbf{x}_i + (\tau \circ \textsf{W}_{ti}\textbf{x}_0) \odot (\textsf{W}_{ai}\textbf{x}_0)) \odot \sigma \circ \textsf{W}_{si}\textbf{x}_0, \, \\ \forall &\textbf{x}_0 \in \mathcal{X}_0 \subset \mathbb{R}^m, \, \forall \textbf{x}_i \in \mathcal{X}_i \subset \mathbb{R}^{m_i}
\end{split}
\end{equation}

Where $\textbf{x}{_0}$ is an ANN input, $\textbf{x}_i$ is an input of the $i^{th}$ layer, $\textsf{W}_i\textbf{x}_i$ is the linear transformation of the main trunk, $\textsf{W}_{ti}\textbf{x}_0, \, \textsf{W}_{ai}\textbf{x}_0, \, \textsf{W}_{si}\textbf{x}_0$ are linear transformations inside the KGate cell, and $\tau, \, \sigma$ are hyperbolic tangent and logistic sigmoid activation functions, respectively. In the non-normalized version, KGate functions as hard GLU, shutting down self-selected channels completely (not worthy of attention), while in the normalized version, it works more similarly to RBF, decreasing the importance of the self-selected channels.

In alignment with Ivakhnenko's approach \cite{ivakhnenko1971polynomial}, the expansion of multi-layer neural network models could be accomplished through the GMDH utilizing a neuron activation function defined by a short-term polynomial. In our specific case, the polynomial corresponds to the second degree of pairwise connected neurons, ensuring the generation of a linear gradient optimization hyperplane at each step of layer generation.

\begin{equation}
\label{eq:9}
f(x_i, x_j) = (w_{ki}x_i + w_{kj}x_j + w_{k0})^2
\end{equation}

The GMDH exhibits the ability to create new layers with enhanced predictive accuracy for novel data. In each layer, the GMDH generates new neurons that are fitted to the training data, and a specific number of the best-fitted neurons are subsequently chosen to form the succeeding layer.

Spectral CNN (SCNN) was organized as parallel Directed Acyclic Graph (DAG) cells, similar to Inception or ResNet type cells \cite{szegedy2015going,ren2016deep}, of $3 \times 1$, $5 \times 1$, $7 \times 1$, $11 \times 1$, $13 \times 1$ dimensions, packets of $16$ of each, \cite{SelitskiyMVSK2022}.

The custom Transformer-type non-linear activation function was implemented in its dot-product version (effectively cosine distance) \cite{VaswaniSPUJGKP17}:

\begin{equation}
\label{eq:10}
\begin{split}
\textbf{z}_i &= (\textsf{W}_{vi}\textbf{x}_i + \textsf{W}_{v0}) \, softmax \circ (\textbf{q}_i^T \, \textbf{k}_i/(d_{ki} + d_{qi}))^T \\
\textbf{k}_i &= (\textsf{W}_{ki}\textbf{x}_i + \textsf{W}_{k0}), \, d_{ki} = mean \circ (\textbf{k}_i^T \, \textbf{k}_i)\\
\textbf{q}_i &= (\textsf{W}_{qi}\textbf{x}_i + \textsf{W}_{q0}), \, d_{qi} = mean \circ (\textbf{q}_i^T \, \textbf{q}_i)\\
\forall &\textbf{x}_i \in \mathcal{X}_i \subset \mathbb{R}^{m_i}
\end{split}
\end{equation}

where $\textbf{x}_i$ is an input, and $\textsf{W}_{v/k/qi}\textbf{x}_i + \textsf{W}_{v/k/q0}$ are linear transformations inside the $i^{th}$ layer.

ANN models were trained using the ``adam'' learning algorithm with $0.01$ initial learning coefficient, mini-batch size $32$, and $1000$ epochs. 

\section{Results}
\label{sec:rs}
As previously indicated, computational experiments were performed on Dow Jones indexes, which were partitioned according to the description in Section~\ref{sec:ds}. Various models, including Linear Auto-regression, ANN Regression, ANN with ReLU, Logistic, Hyperbolic tangent activations, sequence-to-sequence, and sequence-to-value LSTM, as well as simple sequential and spectral cascade CNN, RBF, KGate, and GMDH ANNs, were utilized. These experiments involved traditional non-graph, non-Kinematic Informed baseline ANN architecture \cite{Selitskiy2024Looks}, and the KINN architecture.
Logistic and Hyperbolic tangent activation ANNs, LSTM and GRU, CNN and SCNN, produced similar results; therefore, only one of them is shown Table~\ref{tab:mape.n.1}-Table~\ref{tab:mape.nn.2}.

\begin{table}
\caption{Mean and standard deviation of MAPE over all sessions for regular ANNs and KIANNs. Normalized input, Dow Jones index.}
\label{tab:mape.n.1}
\begin{center}
\begin{tabular}{lcc}
\hline
\multicolumn{1}{l}{\rule{0pt}{12pt}Model} & \multicolumn{1}{l}{Mean MAPE ANN} & \multicolumn{1}{l}{Mean MAPE KINN} \\[2pt]
\hline\rule{0pt}{12pt}
Sig & 0.04849 $\pm$ 0.04227 & 0.04613 $\pm$ 0.04009 \\
Tanh & 0.04867 $\pm$ 0.04182 & 0.04494 $\pm$ 0.03588 \\
LSTM & 0.04744 $\pm$ 0.02811 & 0.04219 $\pm$ 0.02782 \\
RBF & 0.05226 $\pm$ 0.05374 & 0.04700 $\pm$ 0.03625 \\
GMDH & 0.05647 $\pm$ 0.05875 & 0.04750 $\pm$ 0.03458 \\
Transf. & 0.04784 $\pm$ 0.04441 & 0.04493 $\pm$ 0.03947 \\
KGate norm. & 0.04990 $\pm$ 0.04438 & 0.03546 $\pm$ 0.02742 \\[2pt]
\hline
\end{tabular}
\end{center}
\end{table}

\begin{table}
\caption{Wilcoxon signed rank one-side p-value of MAPE distributions similarity over all sessions for regular ANNs and KINNs. Normalized input, Dow Jones index.}
\label{tab:mape.n.2}
\begin{center}
\begin{tabular}{lc}
\hline
\multicolumn{1}{l}{\rule{0pt}{12pt}Model} & \multicolumn{1}{l}{Wilcoxon p-value} \\[2pt]
\hline\rule{0pt}{12pt}
Sig & 0.40700 \\
Tanh & 0.17910 \\
LSTM & 0.00487 \\
RBF & 0.42650 \\
GMDH & 0.09075 \\
Transf. & 0.32580 \\
KGate norm. & 0.00002 \\[2pt]
\hline
\end{tabular}
\end{center}
\end{table}

\begin{table}
\caption{Mean and standard deviation of MAPE over all sessions for regular ANNs and KINNs. Non-normalized input, Dow Jones index.}
\label{tab:mape.nn.1}
\begin{center}
\begin{tabular}{lcc}
\hline
\multicolumn{1}{l}{\rule{0pt}{12pt}Model} & \multicolumn{1}{l}{Mean MAPE ANN} & \multicolumn{1}{l}{Mean MAPE KINN} \\[2pt]
\hline\rule{0pt}{12pt}
KGate & 0.03565 $\pm$ 0.02350 & 0.03546 $\pm$ 0.02742 \\
ANN & 0.03729 $\pm$ 0.02380 & 0.03824 $\pm$ 0.02414 \\
ReLU & 0.03637 $\pm$ 0.02418 & 0.03791 $\pm$ 0.02955 \\
CNN & 0.04055 $\pm$ 0.03109 & 0.04314 $\pm$ 0.03832 \\
SCNN & 0.03684 $\pm$ 0.02435 & 0.03827 $\pm$ 0.02866 \\[2pt]
\hline
\end{tabular}
\end{center}
\end{table}

\begin{table}
\caption{Wilcoxon signed rank one-side p-value of MAPE distributions similarity over all sessions for regular ANNs and KIANNs. Non-normalized input, Dow Jones index.}
\label{tab:mape.nn.2}
\begin{center}
\begin{tabular}{lc}
\hline
\multicolumn{1}{l}{\rule{0pt}{12pt}Model} & \multicolumn{1}{l}{Wilcoxon p-value} \\[2pt]
\hline\rule{0pt}{12pt}
KGate & 0.58880 \\
ANN & 0.72300 \\
ReLU & 0.64840 \\
CNN & 0.50970 \\
SCNN & 0.55420 \\[2pt]
\hline
\end{tabular}
\end{center}
\end{table}

We can even visually see that using KINN architecture noticeably smoothes spurious behaviour of the ANN architectures prone to it; see Figure~\ref{fig:dj.n}. Those architectures include ANN with saturable activations such as sigmoid, polynomials, and exponentials, which are also sensitive to non-normalized data (especially dot-product in Transformer-like architectures). 

The rigorous approach of the statistically significant demonstration of the differences (and, in our case, improvement) of the proposed temporal graph approach of the KINN compared to the traditional ANN, uses range estimation of the accuracy metrics and applies to them non-parametric hypothesis testing algorithm. We use the inferential statistic's Wilcoxon signed rank one-side test shows that distributions of the k-fold validations of the MAPE accuracy metrics are different (definitely better for KINN) with a high confidence level for such architectures LSTM, KGate normalized input, and reasonably high confidence for GMDH and Tanh, see Table~\ref{tab:mape.n.1},  Table~\ref{tab:mape.n.2}.
On the other hand, for those ANN architectures that are not prone to non-normalized data, see Table~\ref{tab:mape.nn.1}, Table~\ref{tab:mape.nn.2}, KINN does not show statistically significant (or at all) improvement.

\section{Conclusions, and Future Work}
\label{sec:con}

We demonstrated that with high statistical significance (for some KINN architectures $>99\%$), such a temporal graphs approach of the Kinematic Informed Artificial Neural Networks of a particular organisation (normalizing LSTM, GDMH, KGate) improves accuracy metrics of the time series predictions, especially in terms of suppressing the spurious prediction behaviour, compared to the non-KINN implementation of the same ANN architectures.

As a bonus result, we can see a need for normalization due to the saturable nature of the activation functions ANNs use; these problematic effects of normalization, which breaks data topology, may be addressed by the use of KINN architecture that enforces preserving neighbourhood proximity during ANN transformation. This shows the main limitation of the algorithm presented in the paper as achieving notable improvement for the case of time series with a single-dimensional target space and saturable non-linear activation functions.

That may point to the future direction of the research - extend KINN mechanisms from this simple one-parametric main stock index prediction auto-regression models to more complex multi-parametric input/output models. Another direction is to use attention memory and sparse training models to tailor them, or their sparse sub-models, to the changing in time process without catastrophic forgetting effects on the past training information.

\bibliographystyle{splncs03.bst}
\bibliography{references.bib}

\end{document}